\documentclass[graybox]{svmult}

\usepackage{mathptmx}       
\usepackage{helvet}         
\usepackage{courier}        
\usepackage{type1cm}        
\usepackage{makeidx}         
\usepackage{graphicx}        
\usepackage{multicol}        
\usepackage[bottom]{footmisc}

\usepackage{amsmath}
\usepackage{algorithm}
\usepackage{algorithmic}
\usepackage{caption}

\newcommand{\BetP}{\mathrm{BetP}}

\newcommand{\Mess}{\mathrm{Mess}}

\DeclareMathOperator{\ocap}{\displaystyle{\small{\textcircled{{\scriptsize $\cap$}}}}}

\makeindex             

\begin{document}
 
\title*{Belief Approach for Social Networks}
\author{Salma Ben Dhaou, Mouloud Kharoune, Arnaud Martin and Boutheina Ben Yaghlane}
\institute{Salma Ben Dhaou \at LARODEC, IHEC \email{salma.bendhaou@hotmail.fr}
\and Mouloud Kharoune, Arnaud Martin \at IRISA, universit\'e de Rennes 1 \email{Mouloud.Kharoune@univ-rennes1.fr}, \\ \email{Arnaud.Martin@univ-rennes1.fr}
\and Boutheina Ben Yaghlane \at LARODEC, IHEC \email{boutheina.yaghlane@ihec.rnu.tn}
}
%
%
\maketitle

\abstract{Nowadays, social networks became essential in information exchange between individuals. Indeed, as users of these networks, we can send messages to other people according to the links connecting us. Moreover, given the large volume of exchanged messages, detecting the true nature of the received message becomes a challenge. For this purpose, it is interesting to consider this new tendency with reasoning under uncertainty by using the theory of belief functions. In this paper, we tried to model a social network as being a network of fusion of information and determine the true nature of the received message in a well-defined node by proposing a new model: the belief social network. }

\section{Introduction}    
\label{sec:1} 
Social networks appeared long before the birth of Internet. A social network can be defined as a group of persons or organizations connected between them by relations and social exchanges which they maintain. However, with the evolution of connection rates and collaborative technologies which are continuously changing, Internet provides access to new networks that are wider, and more playful social but also less easily recognizable.

Furthermore, an important volume of incomplete and imperfect information are spreading on the network.
Therefore, the management of the uncertainty is fundamental in several domains, especially in social networks.
In fact, belief functions theory allows, not only the representation of the partial knowledge, but also the fusion of information.
In the case of social networks, this theory allows to attribute mass functions to the nodes which represent, for example, persons, associations, companies and places as well as links that can be friendly, family and professional and on messages that can be of type for example:  personal commercial, personal not commercial, impersonal commercial and impersonal not commercial.
Therefore, we will have a global view on exchanges made on the network and this will lead us to make a better decision.

In addition, by using uncertainty, we can better monitor the behaviour of the social network. Thus, extending the work on the real plane, we can predict such a terrorist act or assess the quality of a product or follow a buzz\ldots

In this context, previous works have focused on models and methods devoted to the analysis of social network data \cite{Stanley} \cite{Kempe03a} while others have interested in information fusion in order to have a global information about the network \cite{lfa2012_elzoghby}.

The aim of this paper is to propose a new model, a belief social network which is a network supplied by the masses. In fact, we attribute a mass function to the nodes, edges and messages.

This paper is structured as follows. In section 2, we briefly recall some concepts related to the theory of belief functions. We propose in section 3 our model: the belief social network. In section 4, we present the fusion of the masses on belief social network. Finally, section 5 is devoted to illustrate the belief social network and section 6 concludes the paper.

\section{Basic concepts of belief functions}
\label{sec:2}
In this section, we will remind the basic concepts of the theory of belief functions used to instrument our model, the belief social network.  Let $\Omega$ be a finite and exhaustive set whose elements are mutually exclusive, $\Omega$ is called a frame of discernment. A mass function is a mapping $m:2^\Omega \rightarrow [0,1]$ such that $\displaystyle  \sum_{X \in 2^\Omega} m(X)=1$ and $m(\emptyset)=0$. The mass $m(X)$ expresses the amount of belief that is allocated to the subset $X$. In order to deal with the case of the open world where decisions are not exhaustive, Smets~\cite{Smets93a} proposed the conjunctive combination rule.  This rule assumes that all sources are reliable and consistent. Considering two mass functions $m_1$ and $m_2$ for all $A \in 2^{\Omega}$, this rule is defined by:
\begin{equation}
m_{\ocap }(A)= \sum_{B \cap C=A}m_1(B)*m_2(C)
\end{equation}

We will also consider the normalized conjunctive rule, the Dempster rule, given for two mass functions $m_1$ and $m_2$ for all $x \in 2^{\Omega}$ by:

\begin{equation}
m_{\oplus}(A)=\frac{m_{\ocap}(A)}{1-m_{\ocap}(\emptyset)}
\end{equation}

The coarsening corresponds to a grouping together the events of a frame of discernment $\Theta$  to another frame compatible but which is more larger $\Omega$ \cite{Smets93a, Smets97a}. Let $\Omega$ and $\Theta$ be two finite sets. The refinement allows the obtaining of one frame of discernment $\Omega$ from the set $\Theta$ by splitting some or all of its events \cite{Shafer76}.

In order to make a decision, we try to select the most likely hypothesis which may be difficult to realize directly with the basics of the theory of belief functions where mass functions are given not only to singletons but also to subsets of hypothesis. Some solutions exist to ensure the decision making within the theory of belief functions. The best known is the pignistic probability proposed by the Transferable Belief Model (TBM). Other criteria exists like the maximum of credibility and the maximum of plausibility \cite{Janez96a}.

The TBM is based on two level mental models: The ``credal level'' where beliefs are entertained and represented by belief function and the ``pignistic level'' where beliefs are used to make decision and represented by probability functions called the pignistic probabilities. When a decision must be made, beliefs held at the credal level induce a probability measure at the pignistic measure denoted $\BetP$ \cite{Smets90b}. The link between these two functions is achieved by:
\begin{equation}
\BetP(A)= \sum_{B \subseteq \Theta} \frac{|A \cap B|}{|B|} \frac{m(B)}{1-m(\emptyset)}, \forall A \subseteq \Theta
\label{BetP}
\end{equation} 

To focus on the type of relationship between two different frames of discernment, we may use the multi-valued mapping introduced by Hyun Lee \cite{Hyun10a}:
\begin{equation}
m_\Gamma(B_j)=\sum_{\Gamma(e_i)=B_j}m(e_i)
\label{mv}
\end{equation}
with $e_i \in \Omega$ and $B_j \subseteq \Theta$. Therefore the function $\Gamma$ is defined as follow $\Gamma: \Omega \rightarrow 2^{\Theta}$.

The vacuous extension, being a particular case of multi-valued mapping has the objective to transfer the basic belief assignment of a frame of discernment $\Omega$ towards the Cartesian product of frames of discernment $\Omega \times \Theta$. The operation of vacuous extension, noted $\uparrow$, is defined by:

\begin{equation}
m^{\Omega \uparrow \Omega \times \Theta}(B)=
\left\{
\begin{array}{ll}
m^\Omega(A)& \mbox{if } B=A \times \Theta\\
0 &  \mbox{otherwise}
\end{array} 
\right.
\end{equation}

The marginalization allows, from a basic belief assignment defined on a space produced to find the basic belief assignment on one of the frames of discernment of the produced space. This operation, noted $\downarrow$ is defined by:
\begin{equation}
m^{\Omega \times \Theta \downarrow  \Omega}(A)= \sum_{B\subseteq \Omega \times \Theta} m^{\Omega \times \Theta} (B) \quad \forall A \subseteq \Omega
\end{equation}
where $A$ is the result of the projection of B on $\Omega$.

\section{Belief Social Network}
\label{sec:3}
Several works have focused on the representation of networks with graphs. A classical graph is represented by $G=\{V;E\}$ with: $V$ a set of type's nodes and $E$ a set of type's edges. 
This representation does not take into account the uncertainty of the nodes and edges. 

In fact, graphical models combine the graph theory with any theory dealing with uncertainty like probability \cite{Parchas14}, \cite{Khan14} or possibility or theory of belief functions to provide a general framework for an intuitive and a clear graphical representation of real-world problems \cite{Laamari12b}. The propagation of messages in networks has been modelled using the theory of belief functions combined with other theories such as hidden Markov chains \cite{Rammasso09a}. 

\begin{center}
\begin{figure}[h]
\centering
   \includegraphics[scale=0.5]{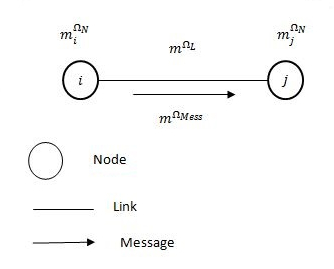}
   \caption{Evidential graph structure for social network.}
\label{EvidentialGraphe}
\end{figure}
\end{center}

In this context, we introduce our model: the belief social network which has the role of representing a social network using the theory of belief functions. Indeed, we will associate to each node, link and message an a priori mass and observe the interaction in the network to determine the mass of the message obtained in a well-defined node.  To do this, we consider an evidential graph $G=\{V^b;E^b \}$ with: $V^b$ a set of nodes and $E^b$ a set of edges. We attribute to every node $i$ of $V^b$ a mass $m_i^{\Omega_N}$ defined on the frame of discernment $\Omega_N$ of the nodes. Moreover, we attribute also to every edge $(i,j)$ of $E^b$ a mass $m_{ij}^{\Omega_L}$ defined on the frame of discernment $\Omega_L$ of the edges. Therefore, we have:
\begin{equation}
V^b=\{V_i,m_i^{\Omega_N}\}
\end{equation} 
and
\begin{equation}
 E^b=\{(V_i^b,V_j^b),m_{ij}^{\Omega_L}\}
\end{equation}
This evidential graph structure is given by Fig~\ref{EvidentialGraphe}. In social network, we can have for example the frame of the nodes given by the classes Person, Company, Association and Place. The frame of discernment of the edges can be Friendly, Professional or Family. Moreover we note: $\Omega_N=\{\omega_{n_1},\ldots,\omega_{n_N}\}$ and $\Omega_L=\{\omega_{l_1},\ldots,\omega_{n_L}\}$.

In social network, many messages can transit in the network. They can be categorized as commercial, personal, and so on.The class of the message is also full of uncertainty. Therefore to each message, we add a mass function in the considered frame of discernment $\Omega_\Mess=\{\omega_{M_1},\ldots ,\omega_{M_k}\}$. 

\section{Fusion of masses on belief social network}
\label{sec:4}

In social network, we can receive the same information from different users. But, can we have the confidence to this information? Moreover, the information can be contradictory. We propose here to take into account the structure of belief social network presented in the previous section to analyse the messages received by one node.

In order to integrate the belief on the nodes and on the edges, we first make a vacuous extension on $\Omega_N \times \Omega_L$ for each mass for the nodes of $V^b$ and on each mass for the edge of $E^b$. Therefore, we obtain on each node $V_i^b$ a mass: $m_i^{\Omega_N \times \Omega_L}$ and on each edge $E_{ij}=(V_i^b,V_j^b)$ between the nodes $V_i^b$ and $V_j^b$ a mass: $m_{ij}^{\Omega_N \times\Omega_L}$. If we consider a coming message from the node $V_i^b$ to the node $V_j^b$ through the edge $E_{ij}$, the belief of the network $m_R^{\Omega_N \times \Omega_L}$ is given by the mass function on the node $V_i$ and the edge $E_{ij}$: 
\begin{equation}
m_R^{\Omega_N \times \Omega_L}=m_{V_i}^{\Omega_N \uparrow \Omega_N \times\Omega_L} \oplus m_{E_{ij}}^{\Omega_L \uparrow \Omega_N \times\Omega_L}
\end{equation}

Here, the index $R$ denotes the resulted belief network from the nodes and the link between them regardless of the message. 

We use the multi-valued operation to combine  mass functions on different frames of discernment. In fact, a multi-valued mapping $\Gamma$ describes a mapping function:
\begin{equation}
\Gamma: \Omega_N \times \Omega_L \rightarrow \Omega_\Mess
\end{equation}

We can calculate these equations by using the formula~\eqref{mv}:
\begin{equation}
\label{massGamma}
\Gamma:m_\Gamma^{\Omega_\Mess}(B_j)=\sum_{\Gamma(e_i)=B_j} m^{\Omega_N \times \Omega_L}(e_i)
\end{equation}
with $e_i \in \Omega_N \times \Omega_L$ and $B_j \subseteq \Omega_\Mess$. From the function $\Gamma$, we can combine the mass given by the network $m_\Gamma^{\Omega_\Mess}$ and the mass of the message to obtain the mass of the message considering the network:
\begin{equation}
\label{m_R}
m_R^{\Omega_\Mess}=m^{\Omega_\Mess} \ocap m_\Gamma^{\Omega_\Mess}
\end{equation}

Now, if we consider $n$ messages coming from $n$ different nodes $V_{i_1}^b, \ldots, V_{i_n}^b$ to the same node $V_j^b$. We can merge the obtained results from the equation~\eqref{m_R} for the $n$ nodes. The goal is to obtain a more precise information on an event describe by the $n$ messages. We then take into account the local network $m_{R_i}^{\Omega_{Mess}}$ of the node $V_j^b$. A local network is defined as a branch of the global network composed of many nodes linked to a same node, the connecting links and the received messages. For example, we can have two nodes which send two messages simultaneously to a third node. We obtain the mass of the global network $m_{G_R}^{\Omega_\Mess}$:
\begin{equation}
m_{G_R}^{\Omega_\Mess}=m_{R_1}^{\Omega_\Mess} \ocap  m_{R_2}^{\Omega_\Mess} \ocap \ldots \ocap m_{R_n}^{\Omega_\Mess}
\end{equation}

Then, we will be able to take a decision on the nature of the resulting message with the pignistic probability using equation~\eqref{BetP}.

\section{Illustrations}
\label{sec:5}
In this section, we will present various experiments conducted to validate our model. We consider three frames of discernment of the nodes, the links and the messages: $\Omega_N=\{Person,Company,Association,Place\}$, $\Omega_L=\{Friendly,Family,Prof.\}$, \linebreak ($Prof.$ for professional), $\Omega_\Mess=\{PC, PNC, IC, INC\}$, with $PC$ for Personal Commercial, $PNC$ for Personal Not Commercial, $IC$ for Impersonal Commercial and $INC$ for Impersonal Not Commercial. We used the passage function $\Gamma$ given in Table~\ref{gamma} which allows.
\begin{table}[ht]
\begin{center}
\begin{tabular}{|c|c|c|c|c|}
  \hline
  $\Gamma$ & Person & Association & Company & Place \\
  \hline
  Friendly & PNC & $PNC \cup INC$ & $PC \cup IC$ & $INC \cup IC$ \\
  Family  & $PNC \cup INC$ & $PNC \cup INC$ & $PC \cup IC$ &$INC \cup IC$  \\
  Professional & $PNC \cup IC$ & IC & IC & IC  \\
  \hline
 \end{tabular}
 \caption{Definition of the function $\Gamma$ given the correspondences between $\Omega_N \times \Omega_L$ and $\Omega_\Mess$.}
  \label{gamma}
\end{center}
\end{table}

For the purposes of our model, we will evaluate three cases. For the first one, we consider a mass function associated to:
\begin{itemize}
 \item a node with: $m^{\Omega_N}(Person)=0.75 \mbox{ and } m^{\Omega_N}(\Omega_N)=0.25$
 \item a link with:  $m^{\Omega_L}(Friendly)=0.75 \mbox{ and } m^{\Omega_L}(\Omega_L)=0.25$
 \item a message with: $m^{\Omega_\Mess}_1(PNC)=0.6 \mbox{ and } m^{\Omega_\Mess}_1(\Omega_\Mess)=0.4$
\end{itemize}
Following our proposed procedure, first, we calculate the vacuous extension of $m^{\Omega_N}$ and $m^{\Omega_L}$ on $\Omega_N \times \Omega_L$ and we combine both mass functions. We obtain:
\begin{eqnarray}
\label{vacuousComb}
\begin{array}{l}
m_R^{\Omega_N \times \Omega_L}(\{Person,Friendly\})=0.5625\\
m_R^{\Omega_N \times \Omega_L}(\{Person, Friendly\},\{Person, Family\},\{Person, Prof.\})= 0.1875\\
m_R^{\Omega_N \times \Omega_L}(\{Person, Friendly\},\{Association, Friendly\},\\
\quad\quad \quad\quad \quad\quad \quad\quad \quad \{Company, Friendly\},\{Place, Friendly\})=0.1875\\
m_R^{\Omega_N \times \Omega_L}(\Omega_N \times \Omega_L)= 0.0625
\end{array}
\end{eqnarray}

Then, we use the $\Gamma$ function to calculate the passage from $\Omega_N \times \Omega_L$ to $\Omega_\Mess$. We obtain:
\begin{eqnarray}
\label{resGamma}
\begin{array}{l}
 m_\Gamma^{\Omega_\Mess}(PNC) =0.5625\\
 m_\Gamma^{\Omega_\Mess}(\Omega_\Mess) =0.4375
 \end{array}
\end{eqnarray}

Then, we make the conjunctive combination of $m_\Gamma^{\Omega_\Mess}$ and $m^\Omega_\Mess$:
\begin{eqnarray}
\begin{array}{l}
m_R^{\Omega_\Mess} (PNC)=0.8250 \\
m_R^{\Omega_\Mess} (\Omega_\Mess)=0.1750
\end{array}
\end{eqnarray}

Finally, to make a decision, we calculate the pignistic probability: 
\begin{eqnarray}
\begin{array}{l}
BetP(PC)=0.0438\\
BetP(IC)=0.0438\\
BetP(PNC)=0.8687\\
BetP(INC)=0.0438 
\end{array}
\end{eqnarray}

If we consider the results, we note that the pignistic probability on Personal Not Commercial is 0.8687. This pignistic probability was equal to 0.7 before considered the network. Hence, we show that considering the network we can reinforce our belief for a given message.

In the second case, we consider the same network, with the same masses $m^{\Omega_N}$ and $m^{\Omega_L}$, but we consider a mass function associated to a message with: $$m^{\Omega_\Mess}_2(PC)=0.6 \mbox{ and } m^{\Omega_\Mess}_2(\Omega_\Mess)=0.4$$
In this case the mass is on the Personal Commercial instead of Personal Non Commercial. As the network is the same we obtain the same mass $m_R^{\Omega_N \times \Omega_L}$ given by equation~\eqref{vacuousComb} as before and also using the $\Gamma$ function the same mass given by the equation~\eqref{resGamma}.

However the result of the conjunctive combination $m_\Gamma^{\Omega_\Mess}$ and $m^{\Omega_\Mess}$ is now:

\begin{eqnarray}
\begin{array}{l}
m_R^{\Omega_\Mess} (\emptyset) =0.3375 \\
m_R^{\Omega_\Mess} (PC) =0.2625 \\
m_R^{\Omega_\Mess} (PNC) =0.2250 \\
m_R^{\Omega_\Mess} (\Omega_\Mess) =0.1750\\
\end{array}
\end{eqnarray}
In this case there is a conflict between the information of the network and the message, therefore a mass come out the empty set. The pignistic probability gives:
\begin{eqnarray}
\begin{array}{l}
BetP(PC)=0.4623 \\
BetP(IC)=0.0660\\
BetP(PNC)=0.4057\\
BetP(INC)=0.0660\\
\end{array}
\end{eqnarray}

We note that in the first example, the highest pignistic probability is associated with the  Personal Not Commercial message that had the larger mass function at the beginning. While in the second example, we find ourselves faced with almost equal probability of Personal Not Commercial and Personal Commercial types  where the need for a second decision on the type of message received.

Now we consider the fusion of the two examples cited above that come on the same node. We obtain the results given in Table~\ref{fusion_res}. We note that by combining the two examples, we get the message Personal Not Commercial that has the highest pignistic probability.

\begin{table}[h!]
\begin{minipage}[t]{.6\linewidth}
\begin{center}

    \begin{tabular}{|c|c|}
 \hline 
Focal & Mass \\ 
 \hline 
$\emptyset$ & $0.5541$ \\ 
 \hline
${PNC}$ & $0.3694$ \\ 
 \hline
${PC}$ & $0.0459$ \\ 
\hline
$\Omega_{Mess}$ & $0.0306$ \\ 
 \hline

 \end{tabular}
\end{center}
\end{minipage}
\hfill
\begin{minipage}[t]{.6\linewidth}
\begin{center}
\hspace{-5cm}
    \begin{tabular}{|l|l|}
 \hline 
Message & BetP \\ 
 \hline 
PC & $0.1202$ \\
\hline
IC & $0.0172$ \\
\hline
PNC & $0.8455$ \\
\hline
INC & $0.0172$\\
\hline
\end{tabular}
\end{center}
\end{minipage}
\caption{Fusion of the two examples: the mass function and the pignistic probability}
\label{fusion_res}
\end{table}

Working on real data, we can assign the mass functions to the nodes, edges and messages by evaluating certain parameters, for example, the type of contacts that are related to the profile in question as well as the type of publications produced (case of facebook). 


\section{Conclusion}
\label{sec:6}
In this work we presented in the first section a general introduction in which we reviewed the notion of social networks and the interest of the proposed method to respond to the expectations for reasoning under uncertainty. In the second section, we briefly introduced the basic concepts used in the theory of belief functions. Then we focused on the introduction of our model and the different notation used. Indeed, we treated step by step development of the construction of the graph.  Finally, we detailed the process of merging the information flowing through the network. We also showed how the process is carried out of the fusion and explained how we can make a decision on the nature of the messages received by using the pignistic probability. In fact, in many cases, we can take a new decision on the nature of the message received by a well-defined node. This idea was explained in the second example in the illustration part. In future work, we aim to represent the update of the elements composing the network as well as to scale.

\bibliographystyle{spmpsci}
\bibliography{biblio2}
\end{document}